\documentclass[12pt]{article}
\usepackage[utf8]{inputenc}
\usepackage{amsmath}
\usepackage{amsfonts}
\usepackage{amssymb}
\usepackage{graphicx}
\usepackage{geometry}
\usepackage{booktabs}
\usepackage{listings}
\usepackage{xcolor}
\usepackage{hyperref}
\usepackage{multirow}
\usepackage{subcaption}
\usepackage{float}

\definecolor{codegreen}{rgb}{0,0.6,0}
\definecolor{codegray}{rgb}{0.5,0.5,0.5}
\definecolor{codepurple}{rgb}{0.58,0,0.82}
\definecolor{backcolour}{rgb}{0.95,0.95,0.92}

\lstdefinestyle{codeStyle}{
    backgroundcolor=\color{backcolour},   
    commentstyle=\color{codegreen},
    keywordstyle=\color{magenta},
    numberstyle=\tiny\color{codegray},
    stringstyle=\color{codepurple},
    basicstyle=\ttfamily\footnotesize,
    breakatwhitespace=false,         
    breaklines=true,                 
    captionpos=b,                    
    keepspaces=true,                 
    numbers=left,                    
    numbersep=5pt,                  
    showspaces=false,                
    showstringspaces=false,
    showtabs=false,                  
    tabsize=2
}
\title{
    \vspace{-1em}
    \rule{\textwidth}{1.2mm} \\[0.5em]
    \textbf{Logarithmic Memory Networks (LMNs): Efficient Long-Range Sequence Modeling for Resource-Constrained Environments} \\[0.5em]
    \rule{\textwidth}{0.4mm}
}

\author{\textbf{Mohamed A. Taha} \\ Independent Researcher \\ \href{mailto:Ahmed.Boin@gmail.com}{Ahmed.Boin@gmail.com}}

\date{\today}

\begin{document}

\maketitle

% Abstract placeholder
\begin{abstract}
Long-range sequence modeling is a crucial aspect of natural language processing and time series analysis. However, traditional models like Recurrent Neural Networks (RNNs) and Transformers suffer from computational and memory inefficiencies, especially when dealing with long sequences. This paper introduces \textbf{Logarithmic Memory Networks (LMNs)}, a novel architecture that leverages a hierarchical logarithmic tree structure to efficiently store and retrieve past information. LMNs dynamically summarize historical context, significantly reducing the memory footprint and computational complexity of attention mechanisms from \(O(n^2)\) to \(O(\log(n))\). The model employs a single-vector, targeted attention mechanism to access stored information, and the memory block construction worker (summarizer) layer operates in two modes: a \textit{parallel execution mode} during training for efficient processing of hierarchical tree structures and a \textit{sequential execution mode} during inference, which acts as a memory management system. It also implicitly encodes positional information, eliminating the need for explicit positional encodings. These features make LMNs a robust and scalable solution for processing long-range sequences in resource-constrained environments, offering practical improvements in efficiency and scalability. The code is publicly available under the MIT License on GitHub: \url{https://github.com/AhmedBoin/LogarithmicMemory}.
\end{abstract}

\textbf{keywords:} Logarithmic Memory Networks, Single-Vector Attention, Hierarchical Structures.

% Sections
\section{Introduction}

The ability to process long sequences is essential for numerous machine learning applications, including natural language processing (NLP), speech recognition, time-series forecasting, and genomics. Efficient handling of long-range dependencies is particularly challenging for traditional sequence models, which often struggle with computational efficiency and memory usage—key considerations for deployment on resource-constrained environments such as mobile or edge devices.
\\

Recurrent Neural Networks (RNNs) \cite{elman1990finding, mikolov2010recurrent, tran2021rnn++, dutta2023neuralmemory}, such as Long Short-Term Memory (LSTM) networks \cite{hochreiter1997long, bahdanau2015neural} and Gated Recurrent Units (GRU) \cite{cho2014learning, cho2014properties}, have historically been the cornerstone of sequence modeling. Despite their ability to capture temporal dependencies through recurrent connections, RNNs face significant limitations, including vanishing or exploding gradients, which hinder their capacity to model very long sequences. Furthermore, the sequential nature of RNN computation limits parallelization, resulting in slow training and inference times.
\\

The advent of Transformers \cite{vaswani2017attention} marked a paradigm shift in sequence modeling, with their self-attention mechanism enabling parallel processing and state-of-the-art performance in tasks like NLP and computer vision. Transformers power models such as BERT \cite{devlin2018bert} and GPT \cite{radford2018improving}, demonstrating remarkable success. However, the quadratic complexity of the self-attention mechanism with respect to sequence length presents significant computational and memory challenges, particularly for long sequences. Variants like Transformer-XL \cite{dai2019transformerxl}, Longformer \cite{beltagy2020longformer}, and Linformer \cite{wang2020linformer} address these issues through mechanisms like recurrence, sparse attention, and low-rank approximations. Yet, these innovations remain computationally intensive and ill-suited for resource-constrained platforms.
\\

Alternative approaches, such as memory-augmented networks, including Neural Turing Machines (NTMs) \cite{graves2014neural} and Memory Networks \cite{weston2014memory}, attempt to address long-range dependencies by leveraging external memory for storage and retrieval. Extensions like the Differentiable Neural Computer (DNC) \cite{graves2016hybrid} improve upon these architectures by introducing dynamic memory allocation and addressing mechanisms. Despite their promise, these models introduce high complexity, training instability, and significant computational overhead, limiting their practicality in large-scale and real-time applications.
\\

State Space Models (SSMs) offer another promising alternative by adopting continuous-time formulations for sequence modeling. Advances such as HiPPO \cite{gu2020hippo}, S4 \cite{gu2021efficiently}, and Mamba \cite{mamba2020} demonstrate the potential of SSMs to handle long-range dependencies with improved stability. However, these models are not without limitations. Training instabilities, such as vanishing or exploding values, and the need for precise weight initialization near 1 to ensure stability remain significant challenges. Additionally, the computational demands of these models often hinder their scalability in real-world applications.
\\

\begin{figure}[ht]
    \centering
    \includegraphics[width=\textwidth]{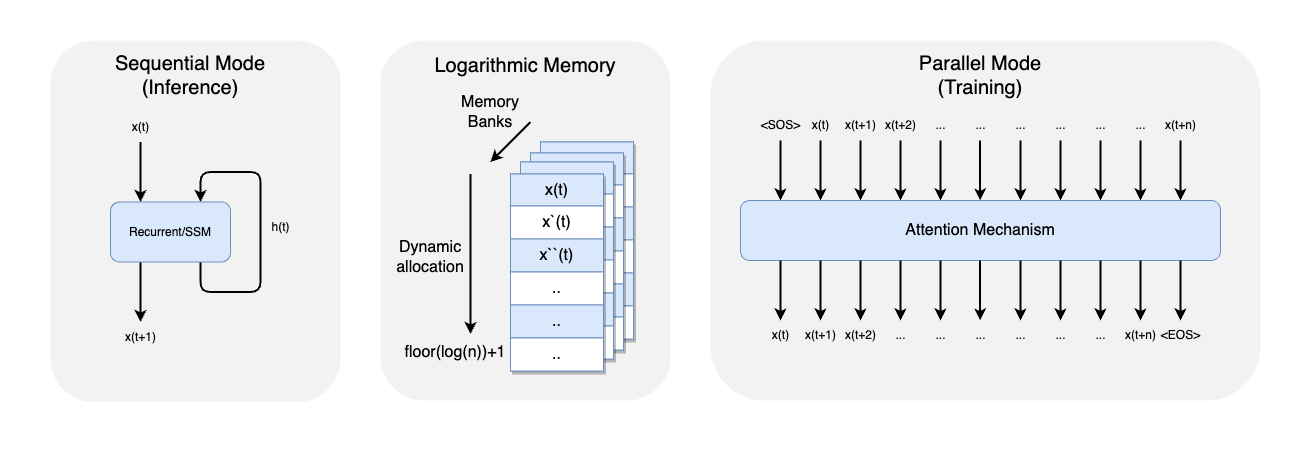}
    \caption{Logarithmic Memory and its capability to train as a transformers and inference as a recurrent}
    \label{fig:LogMem}
\end{figure}

To address these challenges, A \textbf{Logarithmic Memory Networks} is proposed, a novel architecture that introduces a hierarchical logarithmic tree structure for efficient storage and retrieval of long-range dependencies. LMNs leverage a single-vector attention mechanism to dynamically access relevant information from the tree, implicitly encoding positional information and eliminating the need for explicit positional encodings. This design reduces computational complexity and memory usage, making LMNs a practical solution for processing long sequences, particularly in mobile and edge device contexts. Furthermore, LMNs strike a balance between computational efficiency and modeling capacity, offering a robust and scalable approach to sequence modeling. In addition, LMNs feature two distinct modes of operation: a parallel execution mode during training, which enables faster processing, and a sequential execution mode during inference, which acts as a memory management system that significantly reduces the memory footprint. This dual-mode approach enhances the efficiency of LMNs in both training and real-time applications. See Figure \ref{fig:LogMem} for a visual representation of these modes.
\\

This paper is structured as follows: Section \ref{sec:related} reviews related work, Section \ref{sec:lmn} details the LMN architecture, Section \ref{sec:implementation} explains implementation specifics, Section \ref{sec:results} presents experimental findings, Section \ref{sec:discussion} discusses key insights and limitations, and Section \ref{sec:conclusion} concludes with directions for future research.
\section{Related Work}
\label{sec:related}
Several techniques have been developed to overcome the limitations of the existing methods for long sequence modeling. The next section will review some of them here.

\subsection{Recurrent Neural Networks and Their Limitations}
Recurrent Neural Networks (RNNs), including \textbf{Long Short-Term Memory (LSTM)} \cite{hochreiter1997long} and \textbf{Gated Recurrent Units (GRU)} \cite{cho2014learning}, were historically the go-to architectures for processing sequential data. These models attempt to capture temporal dependencies by maintaining a hidden state across time steps. However, RNNs are limited by their inability to effectively capture long-range dependencies due to vanishing and exploding gradients. While LSTMs and GRUs mitigate some of these issues, they still struggle with very long sequences. The complexity of training RNNs, as they require multiple passes through the entire sequence to update the hidden state, leading to slower training times and difficulty in parallelization. This sequential nature further exacerbates the challenge of scaling to large datasets or longer sequences.

\subsection{Memory-Augmented Networks}
To address the inefficiency of RNNs with long sequences, researchers began exploring \textbf{memory-augmented networks} \cite{tan2022hybrid, liu2024learnablememory}. The \textbf{Neural Turing Machine (NTM)} \cite{graves2014neural} and \textbf{Memory Networks} \cite{weston2014memory} are two notable models in this category. NTMs extend the concept of Turing Machines by using an external memory matrix to store and retrieve information, enabling them to model long-term dependencies more effectively than traditional RNNs. Building upon the NTM, the \textbf{Differentiable Neural Computer (DNC)} \cite{graves2016hybrid} introduces additional enhancements, such as dynamic memory allocation and improved addressing mechanisms, further improving memory utilization and scalability. The key complexity of these models arises from the need to access the external memory, which adds a significant computational overhead, especially when scaling to large datasets. Memory retrieval operations, often involving attention mechanisms.

\subsection{Transformers and Their Impact}
\textbf{Transformers} \cite{vaswani2017attention, tay2022synthformer, liu2023recurrentattention, chen2021dynamicmemory, xu2023coformer} revolutionized sequence modeling by introducing a multi-headed self-attention mechanism that allows for the parallel processing of input sequences. Unlike RNNs, Transformers do not rely on sequential processing, making them more efficient for training and inference. However, their reliance on quadratic complexity with respect to sequence length (\(\mathcal{O}(n^2)\)) imposes a significant computational and memory burden, making them inefficient for very long sequences. The self-attention mechanism compares each token to every other token, leading to memory and computation costs that increase rapidly with the sequence length. This makes Transformers particularly challenging when processing long sequences in real-time applications or on devices with limited resources.

To address the limitations of standard Transformers, several Transformer variants have been developed to handle long sequences more efficiently. The key challenge is that the self-attention mechanism in traditional Transformers scales quadratically with the sequence length, which can lead to prohibitively high memory and computational costs for long sequences. Below are some important Transformer variants designed to address this issue:

\begin{itemize}
    \item \textbf{Transformer-XL}: This model introduces recurrence over segments of text, enabling it to maintain memory of previous segments. By retaining information from past segments, Transformer-XL can process longer texts without needing to retrain on past data, making it more efficient for handling long sequences \cite{dai2019transformerxl}.
    \item \textbf{Longformer}: Longformer leverages sparse attention mechanisms, where instead of attending to all tokens in the sequence, it focuses on a smaller subset (e.g., via a sliding window). This reduces both memory and computational costs, allowing the model to process much longer sequences efficiently \cite{beltagy2020longformer}.
    \item \textbf{Linformer}: Linformer takes a similar approach to Longformer by using low-rank approximations for the attention mechanism. This approximation reduces the complexity of attention, enabling the model to handle longer sequences with reduced memory and time requirements \cite{wang2020linformer}.
\end{itemize}

Despite these innovations, the variants still encounter challenges. While they improve the efficiency of long-sequence processing, they do not entirely eliminate the computational cost and memory limitations. These models still struggle to process extremely long sequences in real-time applications or large datasets, especially in scenarios where computational resources are limited.

\subsection{State Space Models (SSMs)}
State Space Models (SSMs) \cite{gu2022scaling, wang2024compactssm} have emerged as a continuous-time approach to sequence modeling. These models treat sequences as states evolving over time and have been shown to improve the stability and computational efficiency of long-sequence processing. Notable works in this area include \textbf{HiPPO} \cite{gu2020hippo}, \textbf{S4} \cite{gu2021efficiently}, and \textbf{Mamba} \cite{mamba2020}, which introduce novel formulations to enhance SSM capabilities. The computational complexity of SSMs is typically linear with respect to the sequence length, \(\mathcal{O}(n)\), which improves upon the quadratic complexity of RNNs and Transformers. However, despite these advancements, SSMs remain challenging to train due to inherent instabilities, such as exploding or vanishing values when processing long sequences. To address this, weights in SSMs often need to be relatively near to 1 for stability; however, even with such constraints, there is no guarantee of stable behavior during training. Additionally, these models require complex optimization techniques and can be computationally expensive in certain scenarios, particularly when parallelized during training.
\section{Methodology (LMNs)}
\label{sec:lmn}
In this section, a detailed explanation of Logarithmic Memory Networks (LMNs), the proposed model, is provided. The core idea of LMNs is the use of a logarithmic tree structure to represent memory. In the following sections, the architecture, the memory construction process, the summarization operation, and the attention mechanism will be discussed.

\subsection{Architecture Overview}

\begin{figure}[H]  % Forces figure to stay in place
    \centering
    \begin{subfigure}{0.6\textwidth}
        \begin{enumerate}
            \item \textbf{Input Embedding}: The input data is passed through an embedding layer, which maps the raw input into a fixed-dimensional vector space, providing a structured representation for further processing.
            \item \textbf{Memory Construction}: The embedded representation is used to construct a memory structure, which is organized in a hierarchical tree format. This allows the model to capture multi-level dependencies across the input.
            \item \textbf{Single Vector Attention}: A single-vector attention mechanism is employed to access the relevant information stored in the hierarchical memory. This mechanism enables the model to focus on important features across long-range sequences.
            \item \textbf{Output Generation}: The results from the attention mechanism are combined with the original input data, generating the final output. This output is a transformed version of the input, considering both short and long-range dependencies.
        \end{enumerate}
    \end{subfigure}\hfill
    \begin{subfigure}{0.4\textwidth}
        \centering
        \includegraphics[width=\textwidth]{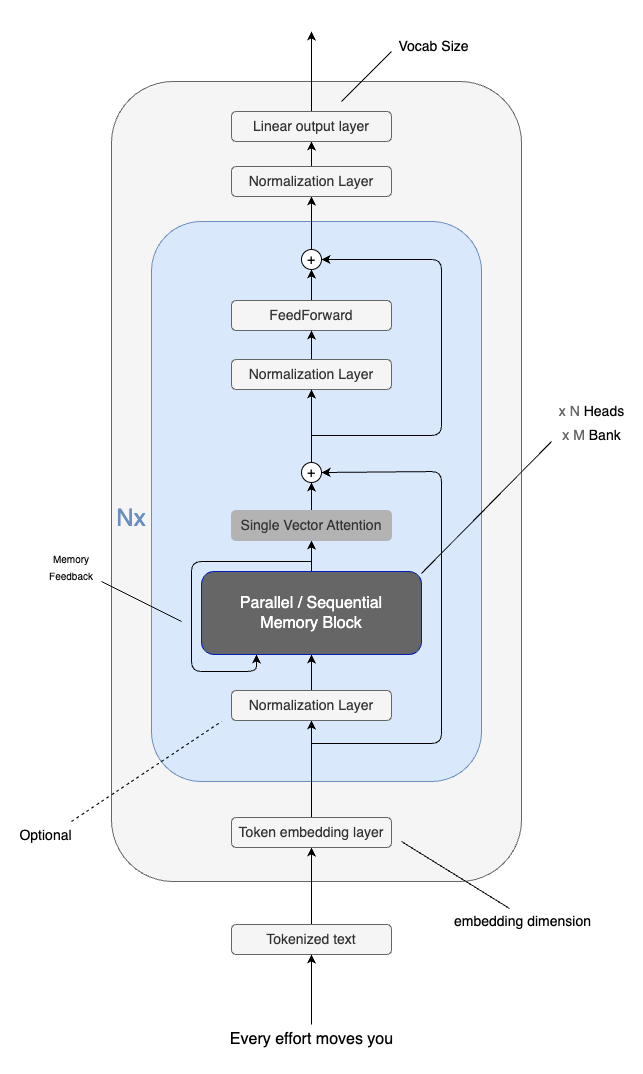}
        \caption{Overview of the LMN architecture.}
        \label{fig:lmn_overview}
    \end{subfigure}
\end{figure}

The Logarithmic Memory Network consists of several key modules designed to efficiently model long-range dependencies through a hierarchical memory structure. The architecture integrates an embedding layer, memory construction, attention mechanism, and output generation, as shown in Figure \ref{fig:lmn_overview}. Each of these modules plays a crucial role in processing and transforming information within the network.

\subsection{Memory Construction}

The memory construction process is crucial for LMN's efficiency. The process constructs the hierarchical tree by summarizing past information, which reduces the memory footprint and provides logarithmic access time to the memory. LMNs provide two ways to perform memory construction: sequential and parallel. The two methods both construct a similar tree-like memory, with the main difference being the process of construction.

\subsection{Summarizer Layer}

The summarizer layer plays a crucial role in condensing information during the memory construction phase. It leverages a linear projection to combine two memory nodes into a single node, effectively summarizing the information from both inputs. Specifically, the layer takes the concatenation of two nodes, represented as a vector of size $2E$ where E is Embedding, and outputs a new node of size $E$ in the subsequent memory location. This output node contains the combined information from the two input nodes, but the memory location is not updated after each operation. Instead, the memory update depends on the position within the sequence.
\\

An important feature of the summarizer layer is its ability to encode relative position information. The input to the summarizer is processed through a linear layer, which inherently captures the features of the nodes along with their relative positions in the sequence. As the memory construction progresses through the sequence, higher-level memory locations begin to encode a more generalized understanding of previous summaries. These summaries retain information about relative positions, which helps in understanding the current position of each summary in relation to its previous context. This enables the model to maintain a coherent sense of the sequence, where each new summary not only integrates prior information but also understands its own relative position within the larger context.
\\

This mechanism facilitates the creation of a hierarchical structure where, in higher-level memory locations, the relative positions of earlier summaries are encoded. This encoding is crucial for maintaining the integrity of the sequence, allowing the model to form a unified concept of the entire sequence by the time it reaches the final memory layer.

\subsubsection{Parallel Memory Construction}
\label{sec:parallel_mem}
\begin{figure}[ht!]
    \centering
    \includegraphics[width=\textwidth]{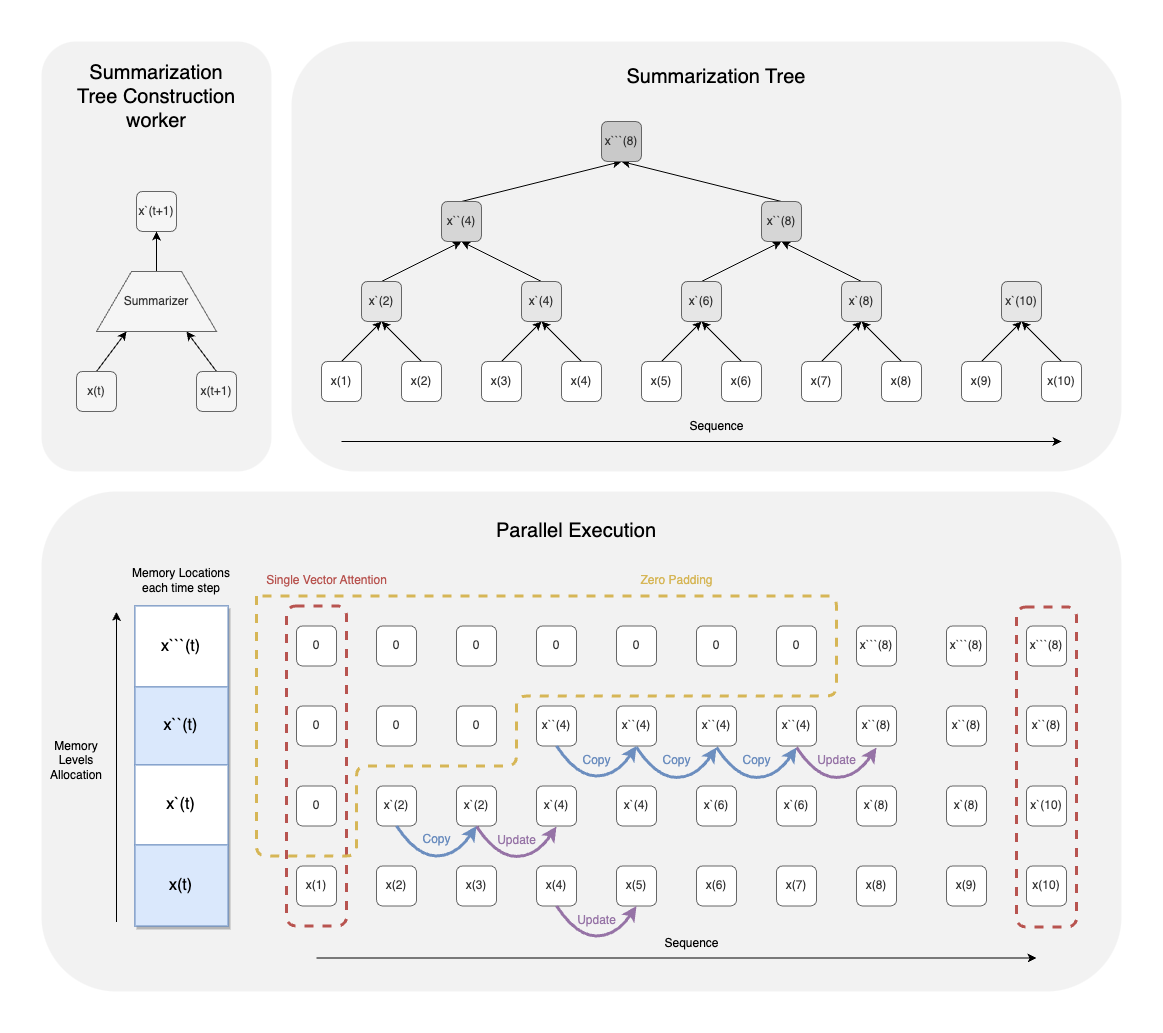}
    \caption{Visualization of parallel memory construction. The nodes are summarized from the bottom up. Nodes from different levels are combined to form the final output.}
    \label{fig:parallel_mem}
\end{figure}

In parallel mode, memory is constructed in a single pass by hierarchically summarizing the past inputs using the `parallel` function, as shown in Figure \ref{fig:parallel_mem}.

\begin{itemize}
    \item \textbf{Initialization}: Input sequence $X$ with a shape of $[B, L, E]$ serves as the lowest level (level 0) of the tree.
    \item \textbf{Iterative Summarization}: In each level $i$, the input sequence of the previous level $x_{i-1}$ is grouped in pairs. These pairs are passed to the summarizer layer to produce a summarized representation for the current level.
    \item \textbf{Expanding and Aligning}: To keep the same size of the input sequence, the generated summarized nodes are repeated to align with the input sequence. The repeating ensures that the number of nodes remain the same for every memory level.
    \item \textbf{Result}: The function returns the memory with the shape $[B, L, \log(L), E]$.
\end{itemize}

\subsubsection{Sequential Memory Construction}
\label{sec:sequential_mem}
\begin{figure}[ht!]
    \centering
    \includegraphics[width=\textwidth]{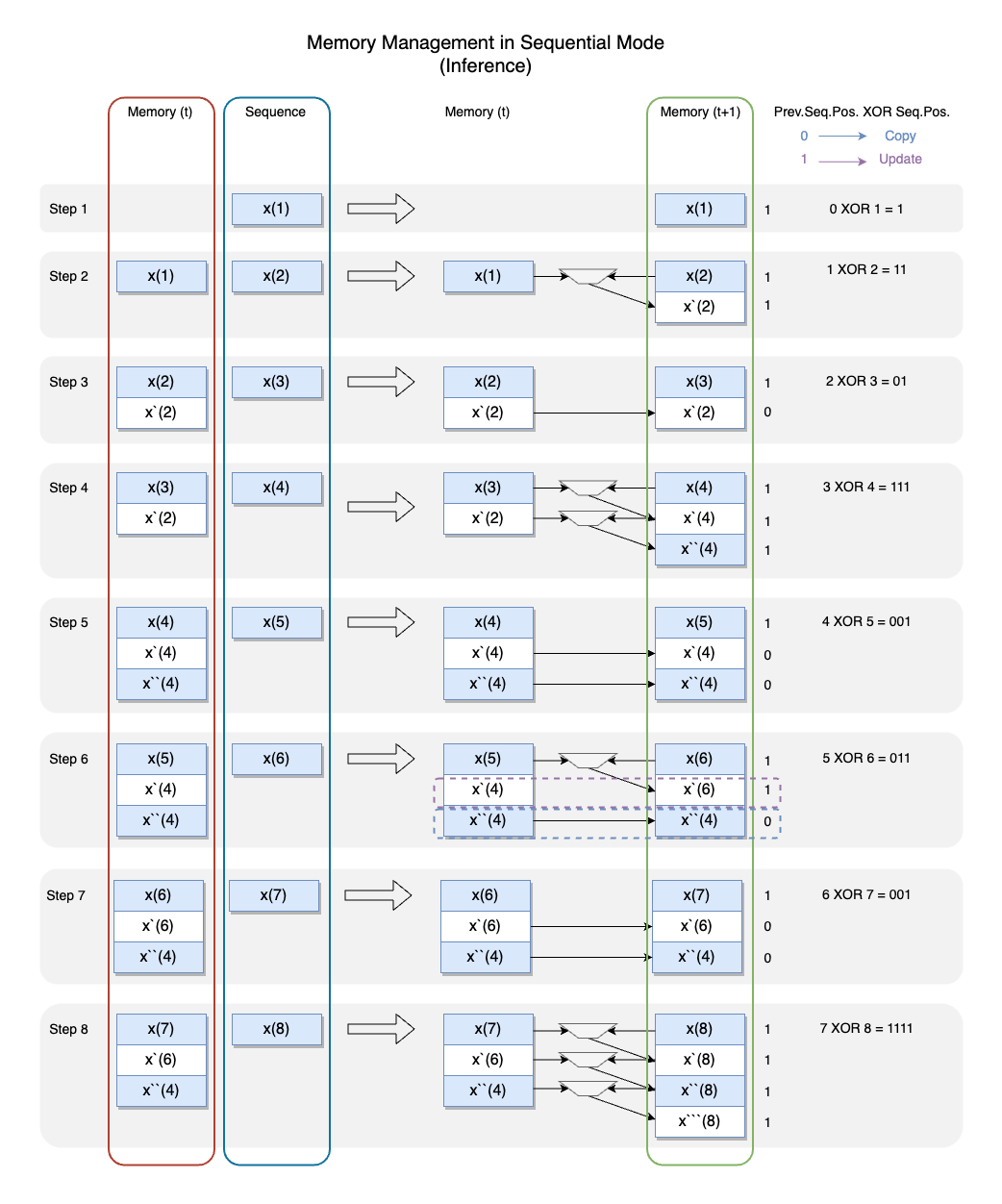}
    \caption{Visualization of sequential memory construction. The pointer keeps track of the input sequence. The bitmask determines whether to summarize or simply a copy of the previous one if no summarization is needed.}
    \label{fig:sequential_mem}
\end{figure}

In sequential construction mode, memory is constructed iteratively, by processing the input sequence sequentially, and summarizing past information when needed. The process is described below and visualized in Figure \ref{fig:sequential_mem}.

\begin{itemize}
    \item \textbf{Initialization}: The initial memory is an empty list. A pointer is used to track the current position within the sequence. It is initialized to 0.
    \item \textbf{Summarization Control}: At each step, the `summarize\_table` method determines whether a summarization should be performed at each level of the memory. The method increments the pointer by one. It generates a bitmask based on the current and previous pointer value. If a bit at position $i$ differs, it means summarization is required at that memory level.
    \item \textbf{Memory Update}: The bitmask determines whether to summarize the existing memory with the current input at each level. If summarization is needed, the `summarizer` layer condenses information and a new memory node is created; otherwise, the old memory node is copied over.
    \item \textbf{Result}: The final memory resembles a hierarchical tree. The final result will be returned in the form of tensor with shape $[B, L, \log(L), E]$, where $B$ is the batch size, $L$ is the sequence length, and $E$ is the embedding dimension.
\end{itemize}

\subsection{Single Vector Attention}

The attention mechanism is a crucial component of LMNs, enabling the model to selectively access past information stored in the logarithmic memory. The single-vector attention mechanism works in the following way, also shown in Figure \ref{fig:attention}:

\begin{figure}[ht!]
    \centering
    \includegraphics[width=\textwidth]{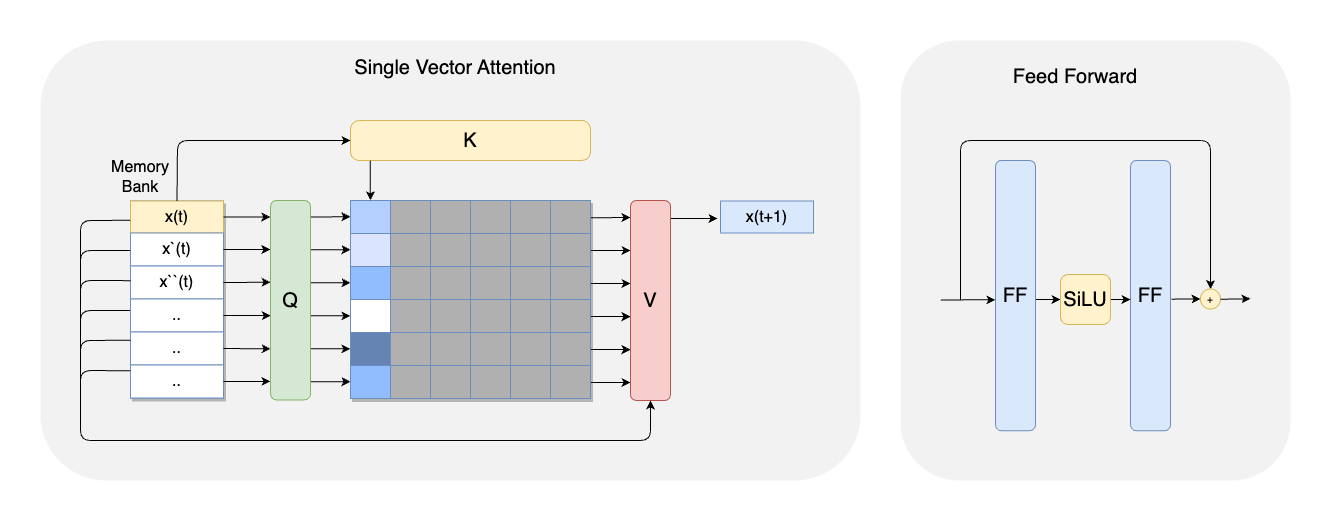}
    \caption{Visualization of the Single Vector Attention mechanism. Q, K and V are generated from the input, and scores are generated by performing matrix multiplication between Q and the transpose of first vector of K, then normalized and masked. Finally, softmax is performed on the result and then weighted sum of the value vector is used to generate the final output}
    \label{fig:attention}
\end{figure}

\begin{itemize}
    \item \textbf{QKV projection}: The memory tensor with shape $[B, L, \log(L), E]$ is passed through the linear layer `qkv` to generate the query (Q), key (K) and value (V). The result is then masked based on the non-zero values of the original input to keep the same zero values if there were any padding if parallelized.
    \item \textbf{Score Calculation}: Attention scores are calculated using the single-vector attention mechanism. The score for each memory level is calculated by performing a matrix multiplication of $Q$ and the transpose of $K$ while only taking the first vector from $K$ which is the current token. The resulting scores have a shape of $[B, L, \log(L)]$. This drastically reduces the complexity of the attention calculation with $\log(sequence)$.
    \item \textbf{Normalization and Masking}: Scores are normalized by dividing the square root of the Embedding $E$ to ensure stable training. The scores are masked by setting non-valid locations to negative infinity to avoid zero padded sequence during softmax.
    \item \textbf{Weighted Sum}: the scores are converted to attention weights using a softmax, which is used to perform a weighted sum of V to produce the output of attention with shape $[B, L, E]$.
    \item \textbf{Reduce Parameters}: The feed forward layer can use the same embedding for both of its layers, effectively reducing the number of parameters. While this approach does not yield the exact same results, it still outperforms GPT-2, achieving better performance with only half the number of parameters.
\end{itemize}

\subsection{Multi-Bank Memory}
\begin{figure}[ht!]
    \centering
    \begin{minipage}[t]{0.6\textwidth} % Adjust the width ratio as needed
        \vspace{10pt} % Remove vertical space at the top
        \raggedright % Align text to the left
        The summarizer layer in each memory bank typically employs a 1D convolution kernel with Depthwise Separable Convolutions to efficiently process the input while reducing the computational cost. Alternatively, a general 1D convolution layer can be used, which may significantly reduce the loss during training. However, this comes at the cost of increased parameters and computational complexity. In some cases, increasing the depth of network blocks can further enhance the effectiveness of the general architecture.
    \end{minipage}%
    \hfill
    \begin{minipage}[t]{0.4\textwidth}
        \vspace{0pt} % Remove vertical space at the top
        \centering
        \includegraphics[width=\textwidth]{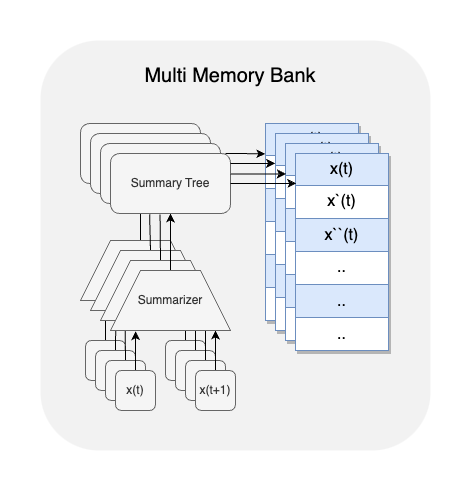}
        \caption{Multi-Memory Banks Visualization.}
        \label{fig:mem_banks}
    \end{minipage}
\end{figure}

To enhance memory performance, multiple memory banks can be utilized to store additional information. In sequential mode or hierarchical tree construction, the summarizer layer is employed to create the memory. Multiple memory banks or trees are generated in parallel using different summarizer layers, with each layer creating its own memory. Subsequently, the single-vector attention mechanism combines all the individual memories into one unified, larger memory, as shown in Fig. \ref{fig:mem_banks}.

\subsection{Path-Through Positional Encoding}

\begin{figure}[ht!]
    \centering
    \includegraphics[width=0.75\textwidth]{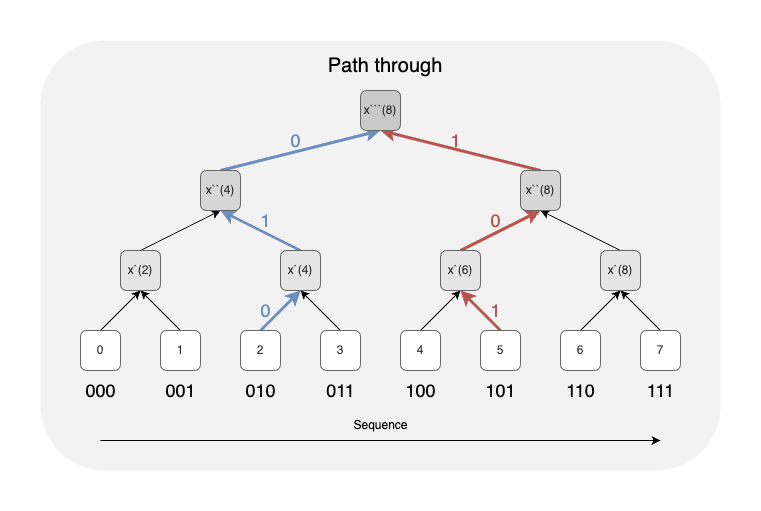}
    \caption{Illustration of relative positional encoding in LMNs. The hierarchical tree structure encodes the relative position of each token as a path through the tree branches or the binary representation of the sequence position minus one. This process occurs during the summarization step, either in parallel mode as the path through the tree or in sequential mode within the memory structure.}
    \label{fig:relative_position}
\end{figure}

Unlike transformers, no explicit positional encoding is added in LMNs. Instead, positional information is implicitly encoded through the hierarchical tree structure. As shown in the figure \ref{fig:relative_position}, the summarization layer encodes relative positions as the paths tokens take during the summarization process. In parallel mode, while in sequential mode, it is derived from the summarization process within the memory structure. This hierarchical encoding effectively embeds positional context corresponds to the path through, eliminating the need for external positional encodings.

\subsection{Expander Summarizer Architecture}
\label{sec:expander_summarizer}

The Expander Summarizer Architecture addresses the summarization bottleneck in long-sequence processing by emulating how humans summarize information. When reading, presenting, or learning, humans initially focus on detailed content but gradually shift to retaining high-level summaries of earlier material. This approach allows them to manage cognitive load, prioritizing new information while maintaining a condensed outline of prior content. For instance, a reader progresses from understanding chapter details to recalling only key ideas, and a teacher emphasizes core concepts over time while condensing earlier material.
\\

The architecture mirrors this process by summarizing older content into hierarchical memory levels while keeping recent information detailed. To mitigate the loss of specific details in heavily summarized distant content, it introduces an expander layer that dynamically increases memory capacity at deeper levels, enabling better retention of critical information. This strategy effectively balances abstraction and detail, much like how the human brain summarizes and recalls information over time.
\\

Although this method closely mimics human memory behavior, it does not fully replicate the ability to retrieve specific details unaffected by summarization. While expanded memory reduces information loss, humans uniquely retain exact details alongside high-level summaries, a capability the model approximates but cannot entirely match. Future efforts aim to develop a mechanism for storing critical information intact, without being affected by the summarization process, to achieve a more complete imitation of human cognitive functions.
\\

\begin{figure}[ht!]
    \centering
    \includegraphics[width=0.85\textwidth]{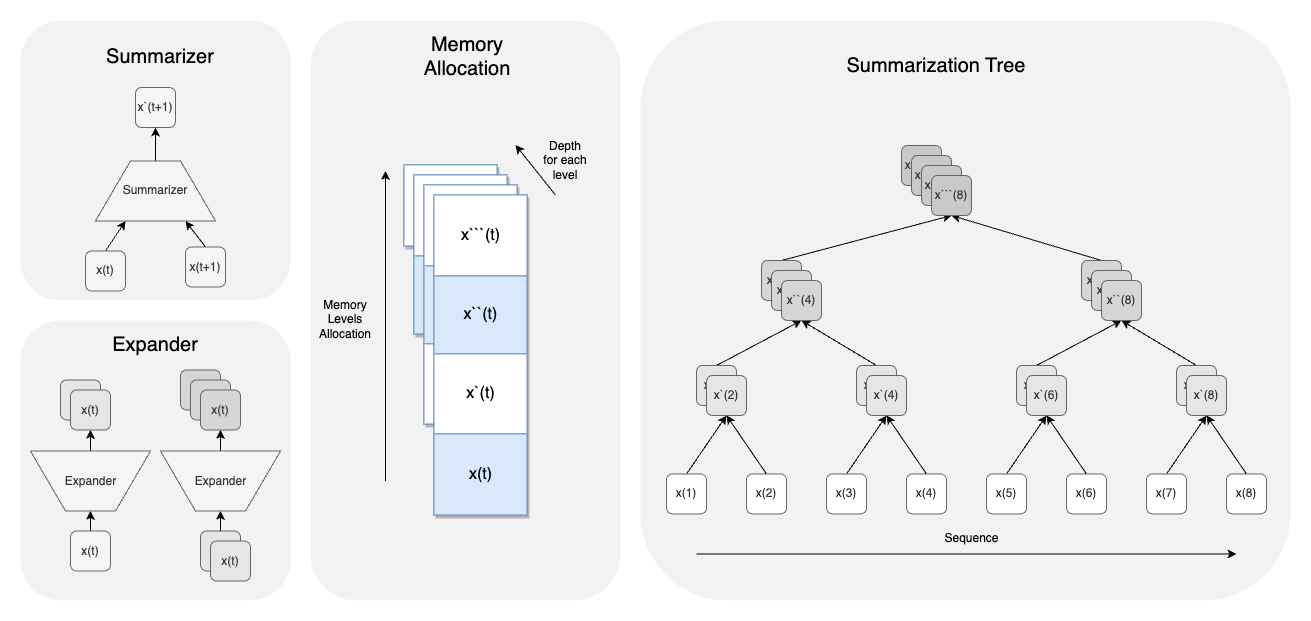}
    \caption{Architecture of the Expander Summarizer. The summarizer (top left) condenses input data into hierarchical memory levels. The expander (bottom left) increases the capacity of each level by adding slots. The center section shows the memory shape at each step, while the right illustrates the tree structure during processing with an expansion factor of 1.}
    \label{fig:expander_summarizer}
\end{figure}

This is achieved using a \textbf{1D Transposed Convolution}, parameterized by an \textit{expansion factor} that determines the number of extra slots added. By default, the expansion factor is set to 1 but can be adjusted as needed to accommodate longer sequences or higher information retention. This mechanism provides flexibility and enhances memory capacity without significantly increasing the network’s width. The process is illustrated in Figure~\ref{fig:expander_summarizer}.

This method enhances scalability while maintaining a hierarchical structure. For a sequence of length $n$ with an expansion factor $k$, the computational complexity is $O(k \cdot \log(n) \cdot (\log(n)-1)/2)$, which simplifies to approximately $O(k/2 \cdot \log^2(n))$ for large $n$. When multiple memory banks are used, the complexity becomes $O(k/2 \cdot \texttt{bank} \cdot \log^2(n))$. This scaling remains efficient compared to the quadratic complexity of traditional attention mechanisms, making it well-suited for extremely long sequences.

Despite these advantages, the increased complexity introduced by the expander layer may pose challenges for resource-constrained devices. While the method is effective for improving information retention and long-sequence processing, its computational and memory demands could limit its applicability in environments with strict resource limitations.
\section{Implementation Details}
\label{sec:implementation}
The implementation of the proposed `LogarithmicMemory` is done using PyTorch and is provided in the repository for transparency and reproducibility. The core implementation is modular, making it easy to integrate into various sequence modeling tasks.

\subsection{Training Process}
During training, the memory is constructed in \textbf{parallel} (described in Section \ref{sec:parallel_mem}). This approach leverages the computational power of GPUs to handle the heavy workload associated with training on very long sequences. Parallel computation ensures efficiency by reducing the time complexity of memory construction during the forward pass.
\\

To enable this, the LogarithmicMemory implementation includes a \texttt{parallelize} flag that can be set to \texttt{True}. When this flag is active, the parallel memory construction method is automatically invoked during the model's forward pass, allowing seamless integration of GPU-accelerated operations into the training pipeline.

\subsection{Inference Process}
Inference is optimized for resource-constrained environments by using the \textbf{sequential memory construction method} (described in Section \ref{sec:sequential_mem}). This method dynamically updates the memory at each step of the inference process, functioning like a recurrent model. The memory storage acts as a feedback hidden state, which makes it highly efficient for scenarios requiring low computational and memory overhead.
\\

During the forward pass of the model in inference mode, the \texttt{sequential} method is invoked, providing a lightweight solution to long-sequence modeling. This enables the system to perform efficiently even on devices with limited resources.

\subsection{Testing and Benchmarking}
The repository includes a suite of \textbf{built-in tests} integrated directly into the core implementation. These tests ensure the functionality, correctness, and performance of the `LogarithmicMemory` module. Additionally, a set of Jupyter notebooks is provided for benchmarking, focusing on inference time and memory footprint for long sequences. Another notebook compares the training and validation loss of `LogarithmicMemory` with attention-based mechanisms (e.g., GPT-2), demonstrating that it achieves competitive or even better results with a significantly smaller number of parameters.
\section{Results}
\label{sec:results}

This section presents an analysis of Logarithmic Memory Networks (LMNs) performance compared to GPT-2 and other models. Evaluations were conducted in two setups:

\begin{itemize}
    \item \textbf{Apple M1 Chip (8-core CPU, 7-core GPU, 8GB unified RAM)}: Computations used the Metal Performance Shaders (MPS) backend in PyTorch.
    \item \textbf{Google Colab (Free Tier with T4 GPU)}: Some experiments were conducted using Nvidia Tesla T4 GPUs.
\end{itemize}

Evaluations focused on parameters, training/validation loss, inference time, computational complexity, and memory usage.

\subsection{Model Comparison: Parameters and Losses}

Table \ref{tab:parameters} compares LMNs with 2 banks, GPT-2, TinyLogMem (Feed-Forward without embedding expansion) with 2 banks and Expander Summarizer Architectures 1 bank \& 1 expander. The same number of memory banks can be used or set to 1 across all models, with efforts made to ensure that all models have nearly similar number of parameters while maintaining comparable performance. The models were trained on the Tiny Shakespeare dataset, with a batch size of 16, block size of 512, and a maximum of 5000 iterations. The learning rate was $1 \times 10^{-3}$, with 200 evaluation iterations.

\begin{table}[ht]
    \centering
    \caption{Comparison of LogMem, GPT-2, TinyLogMem, and ExpSum for 5000 steps}
    \begin{tabular}{|c||c|c|c|c|}
    \hline
    Embedding & Model         & Parameters   & Train Loss       & Val Loss        \\ \hline \hline
    \multirow{4}{*}{32} 
              & GPT-2        & 71,105       & 1.8438           & 1.9704          \\ \cline{2-5} 
              & LogMem       & 71,489       & \textbf{1.5733}  & \textbf{1.7742} \\ \cline{2-5} 
              & TinyLogMem   & \textbf{42,305} & 1.6556        & 1.8540          \\ \cline{2-5} 
              & ExpSum       & 71,489       & 1.6312           & 1.8222          \\ \hline \hline
    \multirow{4}{*}{128} 
              & GPT-2        & 841,281      & 1.3503           & 1.5925          \\ \cline{2-5} 
              & LogMem       & 1,072,193    & \textbf{1.2680}  & 1.5581          \\ \cline{2-5} 
              & TinyLogMem   & \textbf{677,441} & 1.3132        & 1.5699          \\ \cline{2-5} 
              & ExpSum       & 1,072,193    & 1.2981           & \textbf{1.5508} \\ \hline
    \end{tabular}
    \label{tab:parameters}
\end{table}

LMNs have slightly more parameters than GPT-2 but outperform it in both training and validation loss. TinyLogMem, with fewer parameters, still performs competitively. Increasing the embedding size (128) improves Tiny version performance while keeping parameter counts lower than GPT-2.
\\

Further testing with other open-source architectures could be conducted, though this diverges from the primary objective of reducing computational complexity and memory footprint. The results are presented to ensure the model operates as expected and does not exhibit poor performance. Although outperforming existing models was not the focus, the model demonstrates promising results on a smaller scale, with the expectation that such performance may be maintained in larger-scale implementations.

\subsection{Computational Overhead}

LMNs reduce computational complexity from $O(n^2)$ to $O(\log(n))$ with their hierarchical tree structure, The complexity should ideally be \( O(\log(n)^2) \). However, since single-vector attention only needs to focus on the original token, no additional attention is required. This reduces the computational complexity to \( O(\log(n)) \), in the worst case it could be \( O(k/2 \cdot \texttt{bank} \cdot \log^2(n)) \). For example, for $n = 1024$, the compression factor is:

\begin{equation}
    \frac{1024^2}{\log_2(1024)} = 104,857.6 \quad \text{or} \quad 1,048,576\%
\end{equation}

For $n = 8192$, the compression factor becomes:

\begin{equation}
    \frac{8192^2}{\log_2(8192)} = 5,162,220.3
\end{equation}

This reduction becomes more pronounced for longer sequences, making LMNs ideal for long-range dependencies. If QKV values are cached, complexity reduces further to $O(n)$ and $O(\log(n))$.

\subsection{Inference Time and Memory usage}

In attention, it performs similarly to LMNs for short sequences. However, its performance declines significantly as sequence length increases. In sequential mode, we observed performance degradation beyond a sequence length of 2048 with embedding of 32, at which point the processing time became prohibitive. As a result, we shifted to parallel mode at sequence length 2048 and benchmarked up to a maximum of 32,768 due to memory limitations. Traditional attention failed to handle sequences longer than 32,768 due to memory constraints.
\\

\begin{figure}[htbp]
    \centering
    \begin{subfigure}[b]{0.24\textwidth}
        \centering
        \includegraphics[width=\textwidth]{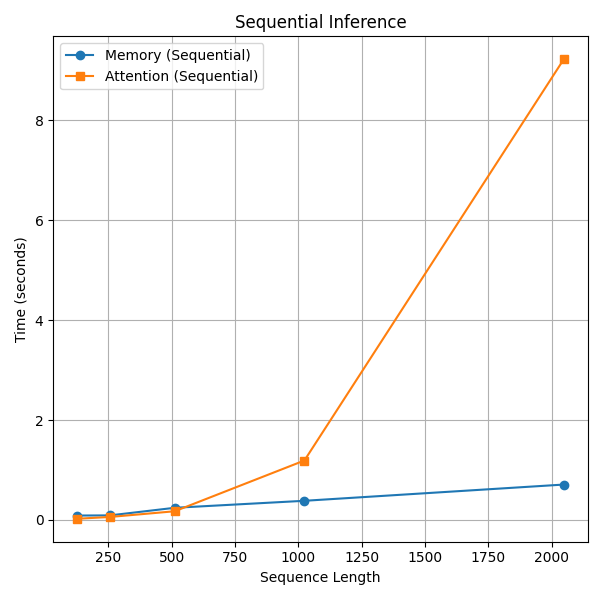}
        \caption{Sequential mode (M1 Chip)}
        \label{fig:time_sequential_M1}
    \end{subfigure}
    \hfill
    \begin{subfigure}[b]{0.24\textwidth}
        \centering
        \includegraphics[width=\textwidth]{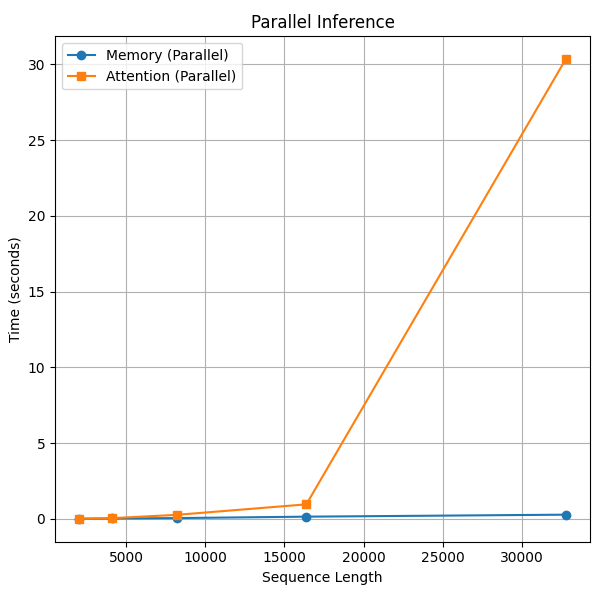}
        \caption{Parallel mode (M1 Chip)}
        \label{fig:time_parallel_M1}
    \end{subfigure}
    \hfill
    \begin{subfigure}[b]{0.24\textwidth}
        \centering
        \includegraphics[width=\textwidth]{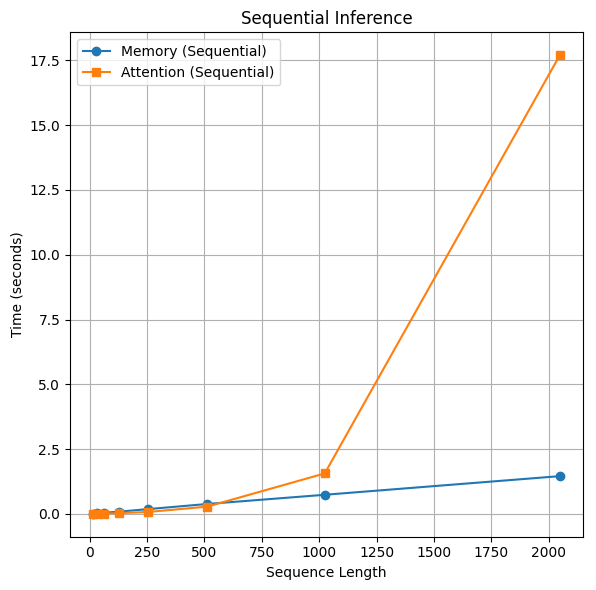}
        \caption{Sequential mode (T4 GPU)}
        \label{fig:time_sequential_T4}
    \end{subfigure}
    \hfill
    \begin{subfigure}[b]{0.24\textwidth}
        \centering
        \includegraphics[width=\textwidth]{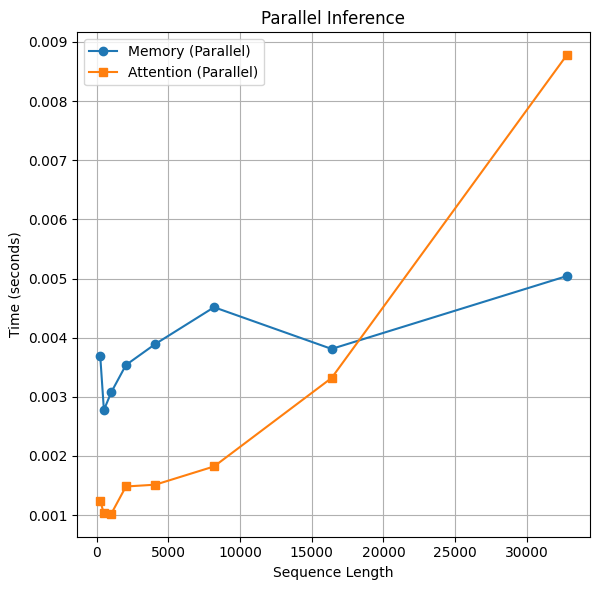}
        \caption{Parallel mode (T4 GPU)}
        \label{fig:time_parallel_T4}
    \end{subfigure}
    \caption{Inference time comparison for Logarithmic Memory and Attention in Sequential and Parallel modes for M1 and T4.}
    \label{fig:time_comparison}
\end{figure}

Figure \ref{fig:time_comparison} presents the inference time comparison between Logarithmic Memory Networks and traditional attention in both sequential and parallel modes.

\begin{figure}[htbp]
    \centering
    \includegraphics[width=0.6\textwidth]{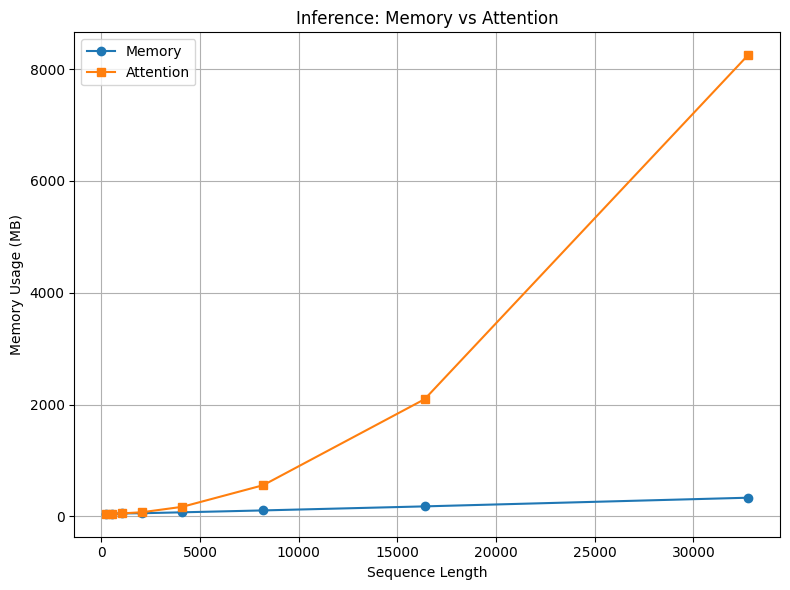}
    \caption{Comparison of memory usage between Logarithmic Memory and Attention. LMNs demonstrate significantly lower memory usage, especially for longer sequences.}
    \label{fig:memory_comparison}
\end{figure}

LMNs demonstrate stable performance across all sequence lengths in both sequential and parallel modes. In sequential mode, LMNs efficiently manage longer sequences, while in parallel mode, they scale without memory issues, unlike traditional attention. Despite memory and time limitations, LMNs outperform traditional attention, particularly for longer sequences, as shown in Figure \ref{fig:memory_comparison}, where LMNs consume significantly less memory, especially for extended sequences. While traditional attention may be more efficient for shorter sequences, LMNs reduce both memory footprint and computational overhead, providing a more resource-efficient solution for longer sequences, making them ideal for resource-constrained environments such as mobile or edge devices.

\subsection{Key Insights}

LMNs achieve competitive results in terms of parameter efficiency and loss performance compared to GPT-2, with their hierarchical memory structure reducing computational overhead and memory usage, particularly for long sequences. LMNs offer higher efficiency without a significant performance sacrifice. They also provide a significant improvement in computational complexity, making them ideal for mobile and edge computing applications, particularly in long-range sequence processing tasks.
\section{Discussion}
\label{sec:discussion}

\subsection{Optimization Challenges and Strategies}
While the LMNs architecture demonstrates considerable potential, its full benefits have not yet been realized due to the need for further optimization. The hierarchical memory structure requires fine-tuning to fully exploit its capabilities. Several modifications have been tested, some producing better results, demonstrating the architecture’s versatility and potential for enhancement across different configurations. These adjustments, such as increasing the depth of the summarization layer, replacing depthwise separable convolutions with standard 1D convolutions, and experimenting with normalization techniques, show significant improvements in model performance. However, they come with trade-offs in computational cost, increased parameter count, and are less suitable for resource-constrained environments. Additionally, approaches like increasing embedding dimensions, expanding memory banks, and scaling multi-head attention can boost capacity, but at the cost of additional computational load.
\\

Interestingly, reducing the feedforward dimension from four times the embedding size to the original size has yielded satisfactory results, lowering both parameter count and computational overhead while maintaining performance. This adjustment proves beneficial in cases where resource efficiency is critical.

\subsection{Improved Memory Management and Model Flexibility}
For potential optimization, extending the summarization layer to summarize more than two tokens at once could reduce the number of memory locations required, leading to a more compact memory representation. Additionally, making the tree structure in the summarization layer more flexible by experimenting with different kernel sizes and varying strides could improve the robustness and adaptability of the architecture, enabling the model to capture complex, hierarchical relationships between tokens. However, increasing the stride of the kernel beyond three means each node in the tree could have more than three branches, requiring the encoding of relative positions of tokens within these branches. In contrast to traditional position encodings, the summarization layer dynamically encodes relative positions based on past summaries, which reduces the need for absolute positional encodings and allows the network to focus on relative dependencies.

\subsection{Challenges in Long Sequence Handling and Future Directions}
Despite these optimizations, large language models still face challenges when dealing with long sequences due to the quadratic growth of the attention mechanism. The inherent quadratic complexity of the attention mechanism becomes a bottleneck for tasks requiring long-range dependencies. The mixture of experts (MoE) \cite{shazeer2017moe, fedus2022switch, kim2023modularmoe} approach can reduce computational overhead by selectively activating subsets of model parameters, but it does not fully address the challenges in processing long sequences. Further optimization is necessary to overcome these challenges.
\\

These optimizations often come with trade-offs, such as higher computational costs and increased memory requirements. However, LMNs show promise beyond the long-sequence problem and could advance language models in applications. With further optimization, LMNs could enable the development of intelligent personal assistants with greater efficiency and contextual awareness, opening doors to smarter, more resource-efficient AI systems for everyday applications in computationally constrained environments.
\section{Conclusion}
\label{sec:conclusion}
This paper introduces Logarithmic Memory Networks (LMNs), a novel architecture designed to address the challenges of long-range sequence modeling in natural language processing and time series analysis. LMNs utilize a hierarchical logarithmic tree structure that efficiently stores and retrieves past information, significantly reducing the memory footprint and computational complexity compared to traditional models like RNNs and Transformers. The model operates in two modes: a parallel execution mode during training to enable faster processing, and a sequential execution mode during inference to optimize memory usage. By incorporating a single-vector attention mechanism and implicitly encoding positional information, LMNs eliminate the need for explicit positional encodings, further reducing computational cost. These features make LMNs a scalable and efficient solution for processing long-range sequences, particularly in resource-constrained environments such as mobile and edge devices. The architecture demonstrates strong potential in advancing sequence modeling tasks and offers a path toward deploying intelligent, real-time AI systems that can operate efficiently, thus enabling more efficient and context-aware AI applications in real-world scenarios.

% Acknowledgment
\section*{Acknowledgment}
I would like to express my deepest gratitude to \textbf{Radwa A. Rakha} for her invaluable support, encouragement, and guidance throughout the development and presentation of this research. Her insightful feedback and unwavering enthusiasm played a pivotal role in refining this work. I am truly grateful for her contributions and dedication, which have greatly enriched the quality of this study.

% Bibliography
\bibliographystyle{IEEEtran}
\bibliography{references}

\end{document}